%% file: main_2025.tex
\documentclass{article}

    \PassOptionsToPackage{numbers, compress}{natbib}

    \usepackage[preprint]{main_2025}

\usepackage[utf8]{inputenc} 
\usepackage[T1]{fontenc}    
\usepackage{hyperref}       
\usepackage{url}            
\usepackage{booktabs}       
\usepackage{amsfonts}       
\usepackage{nicefrac}       
\usepackage{microtype}      
\usepackage{amssymb}
\usepackage{booktabs} 
\usepackage{soul} 
\usepackage{wrapfig}
\usepackage{graphicx}
\usepackage{subcaption}
\usepackage{booktabs} 
\usepackage{multicol}
\usepackage{algorithmic}
\usepackage{algorithm}
\usepackage{amsmath}
\usepackage{amssymb}
\usepackage{amsthm}
\usepackage{graphics}
\usepackage{epsfig}
\usepackage{multirow}
\usepackage{wrapfig}
\usepackage{mathtools}
\usepackage{makecell}
\usepackage{bbm}
\usepackage{booktabs}
\usepackage{graphicx}
\usepackage{kantlipsum}
\usepackage{multicol}
\usepackage{centernot}
\usepackage{xfrac}
\usepackage{afterpage}
\usepackage[dvipsnames]{xcolor}
\usepackage{tablefootnote}
\usepackage{pifont}
\definecolor{mydarkblue}{rgb}{0,0.08,0.45}
\usepackage{hyperref}

\usepackage{amsmath}
\usepackage{amssymb}
\usepackage{mathtools}
\usepackage{amsthm}

\theoremstyle{plain}
\newtheorem{theorem}{Theorem}[section]
\newtheorem{proposition}[theorem]{Proposition}

\theoremstyle{definition}

\theoremstyle{remark}

\newcommand{\SE}{\mathrm{SE}}
\newcommand{\SO}{\mathrm{SO}}

\newcommand{\T}{\mathbb{T}}

\newcommand{\blank}{\rule{0.3cm}{0.25mm}~}

\usepackage{xparse}

\title{EquAct: An SE(3)-Equivariant Multi-Task Transformer for Open-Loop Robotic Manipulation}

\author{%
  Xupeng Zhu \\
  Northeastern University\\
  Boston, MA, 02115 \\
  \And Yu Qi$^*$ \\
  Northeastern University\\
  Boston, MA, 02115 \\
  \And Yizhe Zhu\thanks{Equal contribution.} \\
  Northeastern University\\
  Boston, MA, 02115 \\
  \AND Robin Walters$^\dag$ \\
  Northeastern University\\
  Boston, MA, 02115 \\
  \And Robert Platt\thanks{Equal advising.} \\
  Northeastern University\\
  Boston, MA, 02115 \\
}

\begin{document}

\maketitle

\input{section/0_abstract}
\input{section/1_intro}
\input{section/2_background}

\input{section/3_related_works}

\input{section/4_method}

\input{section/5_experiment}

\input{section/6_conclusion}




\newpage
\bibliographystyle{plainnat}
\bibliography{main_2025}


\newpage
\appendix
\input{section/7_appendix}


\end{document}

%% file: section/0_abstract.tex
\begin{abstract}
Transformer architectures can effectively learn language-conditioned, multi-task 3D open-loop manipulation policies from demonstrations by jointly processing natural language instructions and 3D observations. However, although both the robot policy and language instructions inherently encode rich 3D geometric structures, standard transformers lack built-in guarantees of geometric consistency, often resulting in unpredictable behavior under $\SE(3)$ transformations of the scene. In this paper, we leverage $\SE(3)$ equivariance as a key structural property shared by both policy and language, and propose \textit{EquAct}-a novel $\SE(3)$-equivariant multi-task transformer. EquAct is theoretically guaranteed to be $\SE(3)$ equivariant and consists of two key components: (1) an efficient $\SE(3)$-equivariant point cloud-based U-net with spherical Fourier features for policy reasoning, and (2) $\SE(3)$-invariant Feature-wise Linear Modulation (iFiLM) layers for language conditioning. To evaluate its spatial generalization ability, we benchmark EquAct on 18 RLBench simulation tasks with both $\SE(3)$ and $\SE(2)$ scene perturbations, and on $4$ physical tasks. EquAct performs state-of-the-art across these simulation and physical tasks.
\end{abstract}


%% file: section/1_intro.tex
\section{Introduction}



Recent breakthroughs in multi-task keyframe action policy learning\cite{shridhar2023perceiver, gervet2023act3d, rvt, 3dda, goyal2024rvt, fang2025sam2actintegratingvisualfoundation} have been driven by the success of transformer architectures\cite{vaswani2017attention}, which excel at bridging different modalities—for example, conditioning policies on natural language instructions. Real world manipulation tasks, however,  exhibit rich 3D geometric structure, both in the manipulation policy and the language instructions, which standard transformer architectures do not exploit, e.g., by geometric constraints on their latent features, thereby limiting the geometric consistency of the learned policy. As a result, these multi-task keyframe action methods often fail to generalize to novel 3D scene configurations and require large amounts of robot data to learn the underlying geometric structure from scratch.



This paper presents EquAct, a novel architecture that enforces geometric structure through $\SE(3)$-equivariance in multi-task keyframe action policies and $\SE(3)$-invariance in natural language instructions. We first identify the equivariance inherent to keyframe policies, where the keyframe action adapts to the transformation of the observation, and the invariance of natural language instructions, where given an instruction, the keyframe action transformation depends solely on observation. We then introduce a novel SE(3)-equivariant point transformer U-net with field networks for keyframe action evaluation, alongside novel invariant FiLM (iFiLM) layers to condition the policy on language in a semantically dependent yet geometrically invariant way. 

EquAct enables multi-task keyframe action learning with a simplified, unified model across diverse tasks -- unlike previous $\SE(3)$-equivariant models that allow only for pick and place actions and create separate neural network models for each \cite{simeonov2022neural, ryuequivariant, ryu2024diffusion, huangfourier}.
EquAct is the first method to achieve continuous $\SE(3)$-equivariance (covering both 3D rotation and translation) for multi-task keyframe policy learning, in contrast to\cite{zhu2025coarse, gervet2023act3d, 3dda}, which enforces only translational equivariance. Moreover, most manipulation policy learning works focused on tasks with objects initialized by $\SE(2)$ randomization, despite the pose of objects in the physical world varying in $\SE(3)$. To fill in this gap, EquAct proposes 18 RLBench with $\SE(3)$ initialization to mimic physical world settings. While achieving state-of-the-art performance on $18$ RLBench $\SE(2)$ and $\SE(3)$ benchmarks, our method leverages a spherical Fourier representation to achieve computational efficiency during both training and inference, matching the computation overhead of baselines. Nevertheless, EquAct is limited to keyframe actions that cannot solve fine-grained closed-loop tasks and do not leverage pre-trained vision models.

To summarize, the contributions of this paper are as follows:
\begin{enumerate}
    \item We propose a continuous $\SE(3)$-equivariant keyframe policy that includes a novel equivariant U-net architecture, a novel invariant FiLM layer, and a novel equivariant field network.
    \item We mathematically prove the relevant equivariance and invariance properties.
    \item We validate and find EquAct achieves state-of-the-art performance on 18 RLBench simulation tasks with both $\SE(2)$ and $\SE(3)$ initialization, as well as on $4$ physical tasks.
\end{enumerate}

%% file: section/2_background.tex
\section{Background}

\paragraph{Keyframe imitation learning and multi-task manipulation policy.} 
The keyframe action formulation~\citep{james2022q, james2022coarse} defines an open-loop policy setting, where the policy predicts the next goal pose of the gripper based on the current observation. A motion planner then generates a collision-free trajectory to reach this predicted goal. This formulation decomposes complex trajectories into a sequence of keyframe poses, thereby simplifying policy learning while preserving the ability to solve a wide range of manipulation tasks. Building on this, keyframe imitation learning~\citep{shridhar2023perceiver} formulates the problem as supervised learning, where the policy $\pi(o) = a$ learns to predict the expert keyframe action $a$ given an observation $o$ from expert demonstrations. Specifically, the observation $o = \{s, e\}$ consists of the scene information $s$ and the end-effector state $e = \{e_\text{T}, e_\text{open}\}$, where $e_\text{T} \in \SE(3)$ denotes the gripper pose and $e_\text{open}$ denotes the gripper aperture. Multi-task keyframe manipulation policies~\citep{shridhar2023perceiver, rvt, gervet2023act3d, goyal2024rvt, 3dda} extend this formulation to support multiple skills by conditioning the policy on natural language goals $n$, enabling task-specific behavior across a diverse set of instructions.

\paragraph{Equivariant policy learning.} 
A function $f$ is equivariant with respect to a group $G$ if the group action $g \in G$ commutes with the function, i.e., $f(g \cdot x) = g \cdot f(x)$. In this paper, we focus on the special Euclidean group $\SE(3) = \SO(3) \ltimes \T(3)$, which represents 3D rigid-body transformations composed of 3D rotations $\SO(3)$ and translations $\T(3)$. A group frequently studied in prior work~\citep{wang2021equivariant, zhu2022grasp, huang2023edge, zhao2023integrating,liu2023continual} is the cyclic group $\mathrm{C}_n \subset \SO(2)$, which discretizes the continuous 2D rotation group $\SO(2)$ into $n$ uniformly spaced rotations. An equivariant robotic policy~\citep{pmlr-v164-wang22j, wang2022so2equivariant} satisfies the property:
\begin{equation}
    \pi(g \cdot o) = g \cdot \pi(o), \label{equ:equ}
\end{equation}
meaning the action transforms as the observation transforms. For example, an $\SE(2)$-equivariant planar grasping policy~\citep{zhu2022grasp} predicts a grasp pose from an input image; if the image is rotated, the predicted grasp pose rotates accordingly. There are several strategies to enforce equivariance in neural network-based policies. One common approach is data augmentation~\citep{laskin2020rad, pmlr-v164-wang22j}, where both observations and corresponding actions are transformed according to Equation~\ref{equ:equ} during training. Another method is canonicalization, which transforms the input into a standard reference frame aligned with the action space~\citep{zeng2018learning}. More recently, robot policies that leverage equivariant neural networks~\citep{pmlr-v164-wang22j, zhu2022grasp, huang2023edge,weiler2019general, deng2021vector} have been shown to outperform these alternatives by embedding equivariance directly into the network architecture, but non-of them studies multi-task policy learning.

\paragraph{Spherical harmonics.}
$\SE(3)$-equivariant models rely on feature representations based on spherical functions and spherical harmonics. A spherical function $f_s \colon S^2 \rightarrow \mathbb{R}$ maps a point on the sphere $u\in S^2$ to a real value $y$. An alternative representation of $f_s$ is its Fourier form, where the function is decomposed into spherical harmonic coefficients $c_{l}^m$ via the spherical Fourier transform $\mathcal{F} \colon f_s \mapsto \hat{f}_s$, such that $\hat{f}_s = \{ c_{l}^m \}$. Each coefficient $c_{l}^m$ denotes the weight of the corresponding spherical harmonic $Y_l^m \colon S^2 \rightarrow \mathbb{R}$, which forms an orthonormal basis for the function space $L^2(S^2, \mathbb{R})$. These basis functions are indexed by type (or degree) $l \in \mathbb{Z}_{\geq 0}$ and order $m \in \mathbb{Z}$ such that $-l \leq m \leq l$. The inverse spherical Fourier transform reconstructs the spatial function as $\mathcal{F}^{-1}(f_s)(u) = \sum_{l=0}^{\infty} \sum_{m=-l}^{l} c_l^m Y_l^m(u)$. In practice, truncated spherical coefficients $l\leqslant L_{max}$ are used because they provide a good approximation of the spherical function~\citep{liaoequiformer}. Spherical functions are steerable under $\SO(3)$, making them well-suited for $\SO(3)$-equivariant neural networks~\citep{thomas2018tensor,liaoequiformer,fuchs2020se,passaro2023reducing,liaoequiformerv2}. Specifically, rotating the input function by $g \in \SO(3)$, i.e., $f_s'(u) = g \cdot f_s (u) = f_s(g^{-1}u)$, corresponds to rotating its Fourier coefficients via the Wigner D-matrices $\mathrm{D}$: ${c_l^n}' = \sum_{m} \mathrm{D}^l_{mn}(g) \, c_l^m,$ where ${c_l^n}'$ are the coefficients of the rotated function $f_s'$. For example, a type-$0$ feature is a scalar, and its Wigner D-matrix is identity; a type-$1$ feature is a 3D vector, and its Wigner D-matrix is a 3D rotation matrix.

\paragraph{Spherical CNN.} 
Spherical Convolutional Neural Networks~\citep{cohen2018spherical} lift a spherical function $f_s$ to an $\SO(3)$ function $f_{\SO(3)} \colon \SO(3) \rightarrow \mathbb{R}$ by convolving it with a learnable spherical filter $\psi$ as such $(f_s \star \psi)[g] = \int_{S^2} f_s(u) \psi(g^{-1}\cdot u) \, du, g\in\SO(3).$ This spatial convolution is equivalent to an outer product in the Fourier domain: $\widehat{f_s \star \psi} = \hat{f}_s \cdot \hat{\psi}$, which is more efficient than performing the convolution directly in the spatial domain~\citep{cohen2018spherical, kleeimage}.

%% file: section/3_related_works.tex
\section{Related works}

\paragraph{Keyframe action and multi-task manipulation policy.}







Keyframe action formulation was first introduced by~\citep{james2022q}, which approximates closed-loop manipulator trajectories using a sequence of discrete keyframes, thereby simplifying policy learning. Building on this idea, PerAct~\citep{shridhar2023perceiver} proposes a transformer-based agent that learns a multi-task policy—executing different keyframe actions conditioned on natural language instructions. Later, the multi-task policy learning has diverged into two main directions to evaluate translational action. The first class consists of multi-view-based methods~\citep{rvt,goyal2024rvt,VIHE,zhang2025autoregressive,fang2025sam2actintegratingvisualfoundation}, where the 3D scene is projected into three orthogonal image planes. A ViT-like~\citep{dosovitskiy2020image} multi-view transformer is then used to evaluate translational action values. While this approach is computationally efficient, reasoning in the image plane sacrifices geometric fidelity and requires clever strategies to project into $\SE(3)$~\citep{xu2024se} or $\SO(3)$~\citep{kleeimage,park2022sen} space to achieve $\SE(3)$-equivariance.
The second class operates directly in 3D space~\citep{gervet2023act3d,xian2023chaineddiffuser,3dda,garcia24gembench}, typically using point-cloud-based transformers with densely sampled query points or diffusion models~\citep{chi2023diffusion} to evaluate translational actions. These methods can achieve 3D translational equivariance through 3D CNNs or relative positional embeddings in Transformer~\citep{su2024roformer}, but none incorporate 3D rotation equivariance. Besides translational action, for rotational action prediction, existing approaches typically rely on discretized Euler angles or denoising diffusion over $\SO(3)$ rotations. While the former suffers from gimbal lock and discontinuity issues ~\citep{zhou2019continuity}, the latter incurs significant computational overhead due to iterative refinement. In contrast, \textit{EquAct} achieves both translation and rotation equivariance. It achieves fast inference by evaluating translational and rotational actions in one shot.

\paragraph{Equivariant policy learning.}
Previous works~\citep{van2020mdp, wang2022so2equivariant} have shown that geometric structures are inherent in reinforcement learning problems and that incorporating equivariant policy learning can lead to improved performance. Building on this insight, a series of methods~\citep{zeng2018learning, wang2021equivariant, zhu2022grasp, Huang-RSS-22, wang2022onrobot, zhu2023grasp, liu2023continual, wang2023surprising, nguyen2023equivariant, huang2023leveraging, zhao2023integrating, jia2023seil, huangfourier, kohler2024symmetric, wang2024equivariant, tangri2024equivariant, hu2025pushgrasppolicylearningusing} have proposed $\SE(2)$-equivariant policy learning techniques for robotic tasks. More recently, several methods have extended equivariance to the full $\SE(3)$ group~\citep{simeonov2022neural, ryuequivariant, huang2023edge, ryu2024diffusion, huorbitgrasp, gaoriemann, zhu2025coarse, yang2024equivact, huangimagination, qi2025two, yangequibot, tieseed}. Among these, equivariant closed-loop approaches~\citep{wang2022so2equivariant, jia2023seil, wang2024equivariant, yang2024equivact, yangequibot, tieseed} are typically limited to single-policy learning. Equivariant Pick-and-place methods~\citep{wang2021equivariant, huang2023leveraging, gaoriemann, huangfourier, ryu2024diffusion} assume a fixed action sequence and often require separate models for pick and place actions. Some approaches~\citep{zeng2018learning, wang2021equivariant, huangfourier, liu2023continual, wang2024equivariant, zhu2025coarse, hu2025pushgrasppolicylearningusing} rely on discrete equivariance (e.g., discretized rotations) as an approximation of continuous equivariance. Finally, recent keyframe-based policies that generate poses via denoising~\citep{ryu2024diffusion, 3dda, huangimagination} typically require multiple denoising iterations, resulting in high computational cost. In contrast, \textit{EquAct} learns multi-task keyframe policies using a single unified model, enforces continuous $\SE(3)$ equivariance, and generates keyframe actions in a single, efficient forward pass.

\paragraph{Equivariant neural networks and natural language processing.}
There are several approaches to achieving equivariance in learning-based robotic policies. A common method is data augmentation~\citep{laskin2020rad}, where both inputs and outputs are transformed according to the desired group symmetry during training. Another strategy is canonicalization~\citep{zeng2018learning}, which aligns inputs to a canonical frame prior to inference. 
An alternative is to leverage \textit{equivariant neural networks}, which incorporate equivariance directly into the architecture through symmetry-preserving operations. Prior works~\citep{wang2021equivariant, zhu2022grasp, miller2020relevance} have shown that such networks outperform data augmentation and canonicalization by a significant margin. Equivariant neural networks are grounded in rigorous math from group theory, enabling them to preserve symmetry while maintaining high expressiveness. One class of such networks leverages \textit{group convolutions}~\citep{cohen2016group, weiler2019general, cesa2022program}, which typically discretize a symmetry group and apply convolution over its elements. However, these approaches may suffer from discretization artifacts. Another class operates in the Fourier domain~\citep{geiger2022e3nn, liaoequiformer, passaro2023reducing, liaoequiformerv2}, which offers a more compact and continuous representation of the group. Fourier-based networks allow for continuous equivariance. Our method builds on this Fourier-based framework. Specifically, we scale up EquiformerV2~\citep{liaoequiformerv2} to larger point clouds and to predict full distributions over the $\SE(3)$ action space in multi-task robotic manipulation.

Incorporating natural language into equivariant models has recently gained attention~\citep{li2025large, roche2024equipnas, jia2024open}. \citet{li2025large} and \citet{roche2024equipnas} combine equivariant graph neural networks with invariant language embeddings and evaluate their effectiveness at scale. \citet{jia2024open} introduces a language-conditioned pick-and-place policy using discrete $\SE(2)$ group convolutions. In contrast, our work is the first to explicitly identify the $\SE(3)$ invariance of natural language instructions in the context of robotic policies, and to introduce simple yet effective invariant FiLM (iFiLM) layers to enforce this invariance within an $\SE(3)$-equivariant policy network.

%% file: section/4_method.tex
\section{Method}

\begin{figure}[!t]
    \centering
    \includegraphics[width=\textwidth]{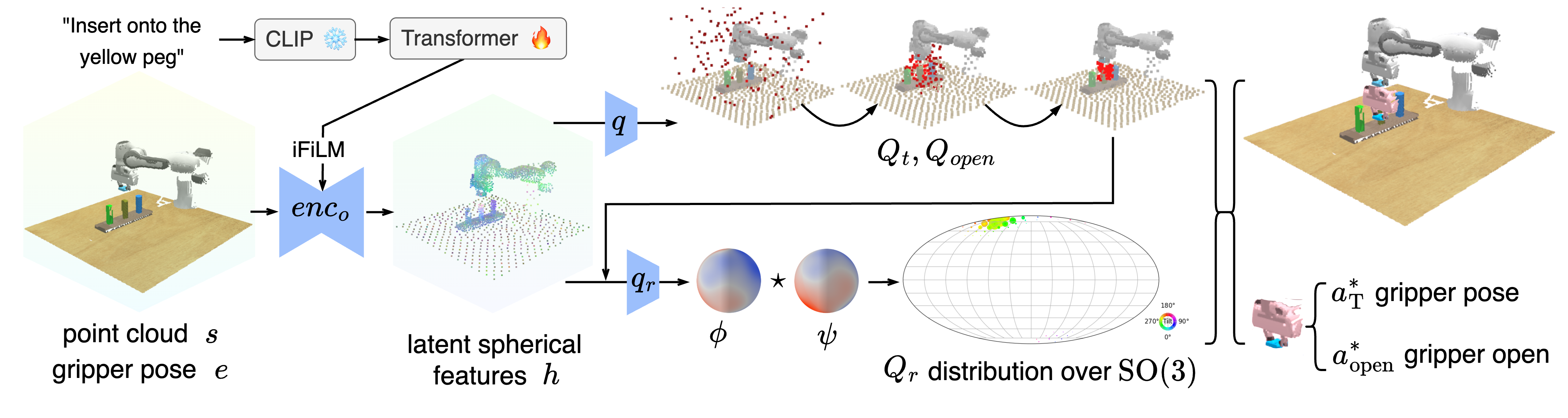}
    
    \caption{\textbf{Overview of EquAct.} EquAct first encodes the observation $o = \{s, e\}$ into latent spherical features $h$ using a $\SE(3)$-equivariant U-Net, $enc_o$, while conditioning the natural language instruction $n$ through invariant iFiLM layers. Based on the encoded features $h$, EquAct then samples and refines translational query actions and gripper open actions using an equivariant field network, resulting in action value functions $Q_t$ and $Q_{\mathrm{open}}$. Finally, a rotational field network aggregates spherical features from $h$ centered at the predicted translation $a_t^*$ to obtain a latent feature $\phi$, which is subsequently convolved with a learned filter $\psi$ to produce the rotational action value function $Q_r$.
}
    \label{fig:equact}
    \vspace{-0.5cm}
\end{figure}


EquAct is a multi-task keyframe action policy that takes an observation $o$ and a natural language instruction $n$ as input and predicts the next best keyframe action of the gripper $a$, denoted by $\pi(o, n) = a$. We model the policy as an implicit function $Q_a(o, n, a) \in \mathbb{R}$ that estimates the action value given an observation and a query action. The inference procedure is illustrated in Figure~\ref{fig:equact} and has the following steps. 
\textbf{1) The $\SE(3)$-Equivariant Point Transformer U-Net} encodes the observation $o$ that includes a point cloud $s$ and the gripper pose $e$ into a set of latent spherical features $h$ at each point in the cloud $h = \mathrm{enc}_o(o)$.
\textbf{2) Invariant Feature-wise Linear Modulation layers} fuse the language embedding $k$, which is treated as type-$0$ features, into the U-Net. Here, $k$ is the encoding of the natural language instruction $n$, by using a CLIP \cite{radford2021learning} tokenlizer and a Transformer \cite{vaswani2017attention} encoder.
\textbf{3) The Equivariant Field Network} takes the latent point cloud $h$ and sampled query actions $a=\{a_t,a_{open},a_r\}$ as input and predicts values for each action $q(a,h)\in\mathbb{R}$. 
The final output actions $a_t^*, a_r^*$ and $a_{\mathrm{open}}^*$ are chosen as those with the highest action values.

During training, EquAct minimizes the following loss:
\begin{align}
    \mathcal{L} &= \mathbb{E}_{(o, n,\bar{a})\sim D, a \sim A}\Big[\mathcal{H}\big(Q_a(o, n, a), \bar{a}\big)\Big] \nonumber\\
    &= \mathbb{E}\Big[\mathcal{H}(\sum_{i=1}^3 Q_{t}(a_t^i, o, n), \bar{a}_t) + \mathcal{H}(Q_r(a_r, \bar{a}_t, o, n), \bar{a}_r) + \mathcal{H}(Q_{open}(a_{open}, \bar{a}_t, o, n), \bar{a}_{open})\Big],
    \nonumber
\end{align}
where $(o, n, \bar{a}) \sim D$ are expert demonstrations consisting of observations $o$, natural language instructions $n$, and expert actions $\bar{a}$, and $a \sim A$ denotes a sampled query action from the action space. $\mathcal{H}$ denotes cross-entropy loss. Intuitively, this loss treat policy learning as a classification problem in which the goal is the policy to correctly choose the expert action from among all available actions.
During training, we also augment the dataset with respect to equation \ref{equ:equ} by randomly rotating the point cloud and the action simultaneously with $[\pm5^\circ, \pm5^\circ, \pm45^\circ]$ rotation along $[x,y,z]$ axis.

\subsection{Equivariance assumptions in multi-task manipulation policy learning}

\begin{wrapfigure}{r}{0.45\textwidth}
    \centering
    \includegraphics[width=0.4\textwidth]{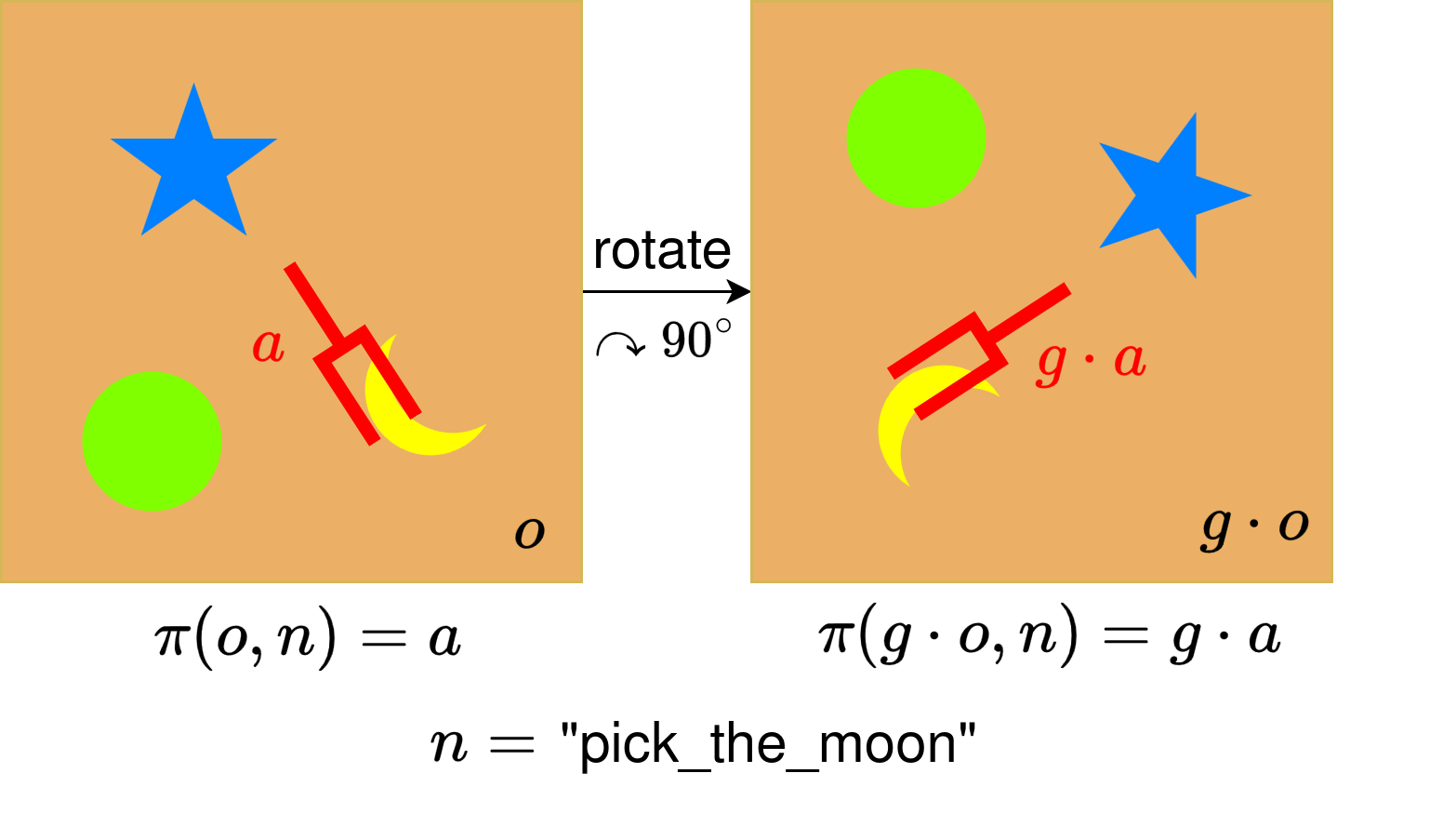}
    \caption{\textbf{The equivariance and invariance of the multi-task keyframe policy.} Under the equivariance assumption, when the observation is transformed to $g \cdot o$, the predicted action transforms accordingly to $g \cdot a$. Under the invariance assumption, given a fixed natural language instruction $n$, the action transformation depends solely on the transformation applied to the observation.}
    \label{fig:equ_inv}
    \vspace{0.1cm}
\end{wrapfigure}

EquAct assumes that the keyframe action policy is equivariant with respect to the observation. That is, when the observation undergoes a transformation, the predicted action should transform accordingly \cite{wang2021equivariant, zhu2022grasp, Huang-RSS-22, ryuequivariant}. Additionally, we identify and assume that the action is invariant to the natural language instruction—meaning that for a fixed instruction, the action should transform solely based on $\SE(3)$ transformation of the observation. Formally, this behavior is expressed as:
\begin{equation}
    \pi(g \cdot o, n) = g \cdot a, \quad g \in \SE(3),
    \label{equ:equact}
\end{equation}
where the group action $g$ operates on both the observation $o$ and the predicted action $a$ by applying rigid-body transformations to the point cloud $s$ or the gripper poses $e_\text{T}$ and $a_\text{T}$, see Figure ~\ref{fig:equ_inv} for an illustration.

Methodologically, EquAct achieves equivariance between the observation and action by employing a novel $\SE(3)$-equivariant Point Transformer U-Net (Section~\ref{sec:EPTU}) and $\SE(3)$-equivariant field networks (Section~\ref{sec:field_nn}). In parallel, it enforces invariance with respect to natural language instructions via the proposed $\SE(3)$-invariant layer-wise modulation (iFiLM) layers (Section~\ref{sec:iFiLM}).

\begin{proposition}
\label{proposition:equact}
    EquAct is $\SE(3)$-equivariant in observation-action mapping and $\SE(3)$-invariant to nature language instruction, as described in Equation ~\ref{equ:equact}.    
\end{proposition}

This is proved by induction; see Appendix \ref{proof:equact}.

\subsection{Equivariant Point Transformer U-net (EPTU)}
\label{sec:EPTU}

\begin{figure}[!ht]
    \centering
    \begin{subfigure}[b]{\textwidth}
        \centering
        \includegraphics[width=\textwidth]{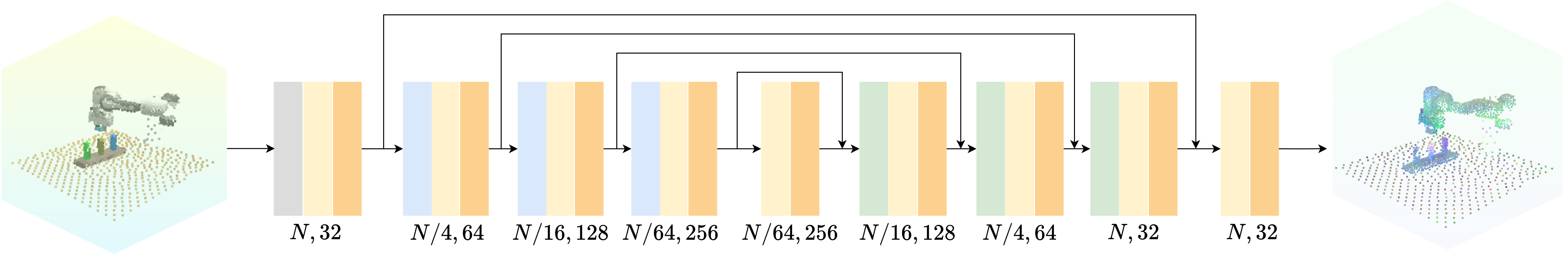}
        \caption{Overview of EPTU.}
    \end{subfigure}
    \begin{subfigure}[b]{\textwidth}
    \centering
    \includegraphics[width=0.85\textwidth]{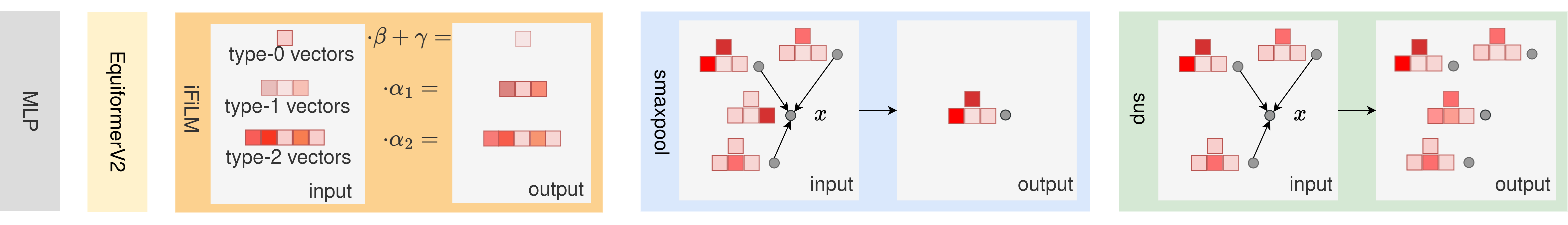}
    \caption{Detailed structure design for each module. Red color indicates the magnitude of the feature.}
    \end{subfigure}
    \caption{\textbf{$\SE(3)$-Equivariant Point Transformer U-net (EPTU).}}
    \label{fig:eptu}
\end{figure}

The $\SE(3)$-equivariant Point Transformer U-Net (EPTU, Figure \ref{fig:eptu}) encodes a point cloud $s$ into equivariant latent features by propagating both local and global information across points. Compared to non-equivariant counterparts such as Point Transformer \cite{zhao2021point} and U-Net \cite{ronneberger2015u}, EPTU achieves continuous $\SE(3)$-equivariance by leveraging spherical Fourier features in its hidden layers. EPTU further improves the computational efficiency of EquiformerV2 \cite{liaoequiformerv2} by adopting a U-net-style architecture \cite{ronneberger2015u}, which incorporates novel \textit{spherical Fourier maxpooling} layers to compress point cloud features and \textit{spherical Fourier upsampling} layers to reconstruct features back to the original resolution. These pooling and upsampling layers are interleaved with standard EquiformerV2 \cite{liaoequiformerv2} graph attention blocks, which first construct a $k$-nearest-neighbor graph for each point and then apply equivariant attention-based message passing. EPTU also incorporates skip connections \cite{ronneberger2015u} between the downsampling and upsampling stages. Compared to prior equivariant point U-Net \cite{ryu2024diffusion, huorbitgrasp}, the proposed U-Net is straightforward to implement, by eliminating the need for caching graphs, i.e., the sup block in Figure \ref{fig:eptu} (b) does not need the maxpool graph in the smaxpool model.

\paragraph{Spherical Fourier maxpooling.}
Analogous to the maxpooling operation in convolutional neural networks \cite{lecun1998gradient}, the \textit{spherical Fourier maxpooling} layer (Figure \ref{fig:eptu} (b) middle) reduces the resolution of the feature map in the spherical Fourier domain. Specifically, for a point $x$, the layer aggregates features from its $k$-nearest neighborhood $\{ c_{l,p} \mid p \in knn(x) \}$ and selects the spherical Fourier coefficient with the largest magnitude at each degree $l$:
\begin{equation}
    c_{l,x}' = \text{smaxpool}\{ c_{l,p} \mid p \in knn(x) \} = c_{l,p^*}, \quad p^* = {\arg\max}_{p \in knn(x)} \| c_{l,p} \|_2^2,
    \label{equ:sfm}
\end{equation}
where the $2l+1$-dimension vector $c_{l,p} = [c_{l,p}^{-l}, c_{l,p}^{-l+1}, \ldots, c_{l,p}^l]$ denotes the type-$l$ spherical Fourier coefficient at point $p$.

\begin{proposition}
\label{proposition:smaxpool}
The spherical Fourier maxpooling operation defined in Equation~\ref{equ:sfm} is $\SE(3)$-equivariant. That is, for any $r \in \SO(3)$ and $t \in \T(3)$:
\begin{equation}
    \mathrm{D}(r) \cdot c_{l,\, t + x}' = \text{smaxpool}\{ \mathrm{D}(r) \cdot c_{l,p} | p\in t + knn(x) \}.
    \label{pro:smaxpool}
\end{equation}
\end{proposition}

This is proved by the orthogonal property of Wigner D-matrices, see Appendix \ref{proof:smaxpool} for a proof.

\paragraph{Spherical Fourier upsampling.}
Interpolation is commonly used for upsampling feature maps \cite{zhao2021point, ronneberger2015u}. To extend this operation to the spherical Fourier domain, we propose a novel \textit{spherical Fourier upsampling} method (Figure \ref{fig:eptu} (b) right). Specifically, for each type-$l$ component, we perform a coefficient-wise interpolation over the $k$-nearest neighbors of a query coordinate $x$:
\begin{equation}
    c_{l,x}' = \text{sup}\{c_{l,p}, x | p\in\text{knn}(x)\} = \text{softmax}_{p\in\text{knn}}\left( \frac{1}{\|x - p\|} \right) c_{l,p},
    \label{equ:sup}
\end{equation}

\begin{proposition}
\label{proposition:sup}
The spherical Fourier upsampling operation defined in Equation~\ref{equ:sup} is $\SE(3)$-equivariant. Specifically, for any $r \in \SO(3)$ and $t \in \T(3)$:
\begin{equation}
    \mathrm{D}(r) \cdot c_{l,\, t + x}' = \text{sup}\{\mathrm{D}(r) \cdot c_{l,\, p},\, t + x | p\in t + \text{knn}(x)\}.
    \label{pro:sup}
\end{equation}
\end{proposition}

See Appendix \ref{proof:sup} for a proof. The proof is based on Schur's lemma \cite{schur1905neue} and that the linear $\SO(3)$ action on the Fourier coefficients.

\subsection{Invariant Feature-wise Linear Modulation Layers (iFiLM)}
\label{sec:iFiLM}
We propose invariant Feature-wise Linear Modulation (iFiLM) layers (Figure \ref{fig:eptu} (b) left) to enforce the geometric invariance of natural language conditioning in the policy, as defined in Equation~\ref{equ:equact}. Unlike standard FiLM layers \cite{perez2018film}, which do not guarantee equivariance or invariance, the iFiLM layer is provably $\SE(3)$ invariant with respect to the conditioning input $k$. Specifically, the iFiLM layer takes as input a spherical Fourier feature $c$ and a type-$0$ (invariant) condition feature $k$, and outputs a semantically modulated feature $c'$:
\begin{align}
    c' = \text{iFiLM}&(c, k), \quad \alpha_l, \beta, \gamma = \text{MLP}(k),\\
    c_l' &= \alpha_l c_l, \quad \text{for } l > 0, \label{line:l>0}\\
    c_0' &= \beta c_0 + \gamma, \quad \text{for } l = 0, 
\end{align}
where iFiLM first uses a multi-layer perceptron to project the condition $k$ into type-$0$ modulation scales $\alpha, \beta$ and bias $\gamma$. Then, iFiLM scales the type-$l$ input feature $c_l$ by $\alpha_l$ for all $l > 0$, and applies an affine transformation to the type-$0$ features using $\beta$ and $\gamma$.

\begin{proposition}
\label{proposition:iFiLM}
The invariant feature-wise linear modulation (iFiLM) layer is $\SO(3)$-invariant with respect to the condition input $k$, and $\SO(3)$-equivariant with respect to the input feature $c$. Specifically, for any rotation $r \in \SO(3)$:
\begin{equation}
    \mathrm{D}(r) \cdot c' = \mathrm{iFiLM}(\mathrm{D}(r) \cdot c,\, k).
\end{equation}
\end{proposition}

See Appendix \ref{proof:iFiLM} for a proof. The proof utilizes Shur's lemma  \cite{schur1905neue}.

\subsection{Equivariant field network}
\label{sec:field_nn}

EquAct evaluates actions in the entire pose action space $A_\mathrm{T}\subset \SE(3)$, rather than action at each point of the point cloud. Therefore, EquAct proposes equivariant field networks $q$ to propagate features from the latent point cloud $h$ to any query point $a_\mathrm{T}\in A_\mathrm{T}$.

For translational action value evaluation, given the query translational action $a_t$ and the latent point cloud $h$, the field network $q_t$ builds a graph with $h$ as the source and $a_t$ as the destination, then performs graph attention to aggregate spherical Fourier features from $h$ to $a_t$. The graph connects the query point to the $k$-nearest neighbor in $h$. The graph attention is implemented by one EquiformerV2 attention block \cite{liaoequiformerv2}. The graph building and attention operation is similar to \cite{gervet2023act3d,ryu2024diffusion,chatzipantazismathrm}, except that the output, i.e., the translation action value is invariant to rotation \cite{wang2021equivariant,zhu2022grasp}: $q_t(a_t, h) = q_t(a_t, g\cdot h), g\in\SO(3)$. Therefore, the field network only takes the type-$0$ feature from aggregated features. We evaluate the translational action in a coarse-to-fine fashion, where the initial resolution of action is coarse, and the subsequent sampling refines the action. The gripper open action $q_{open}(a_{open}, a_t, h)$ is evaluated in the same way, except that $q_{open}$ outputs two channels of type-$0$ features, corresponding to the open/close action values.


For rotational action value evaluation, given the query trans-rotal action $a_t, a_r$ and the latent point cloud $h$, the field network $q_r$ first aggregates features in the same way as the translational network to obtain the spherical Fourier features $\hat{\phi}$ at $a_t$. Then the action value for $a_r$ is calculated by a spherical CNN \cite{cohen2018spherical} with a learnable filter $\hat{\psi}$: $q_r(a_r, a_t, h) = (\phi\star\psi)[a_r] = \mathcal{F}^{-1}(\hat{\phi}\cdot\hat{\psi})[a_r]$. Notice that our field network $q_r$ performs spherical convolution at a 3D location $a_t$ and is $\SE(3)$ equivariant, which differs from the previous $\SO(3)$ equivariant spherical convolution  \cite{cohen2018spherical, kleeimage, howell2023equivariant} that operates in images.

%% file: section/5_experiment.tex
\section{Experiments}

\subsection{Simulation experiments}
\label{sec:sim_exp}

\begin{figure}[!ht]
    \centering
    \includegraphics[width=0.95\textwidth, trim=15mm 100mm 65mm 70mm, clip]{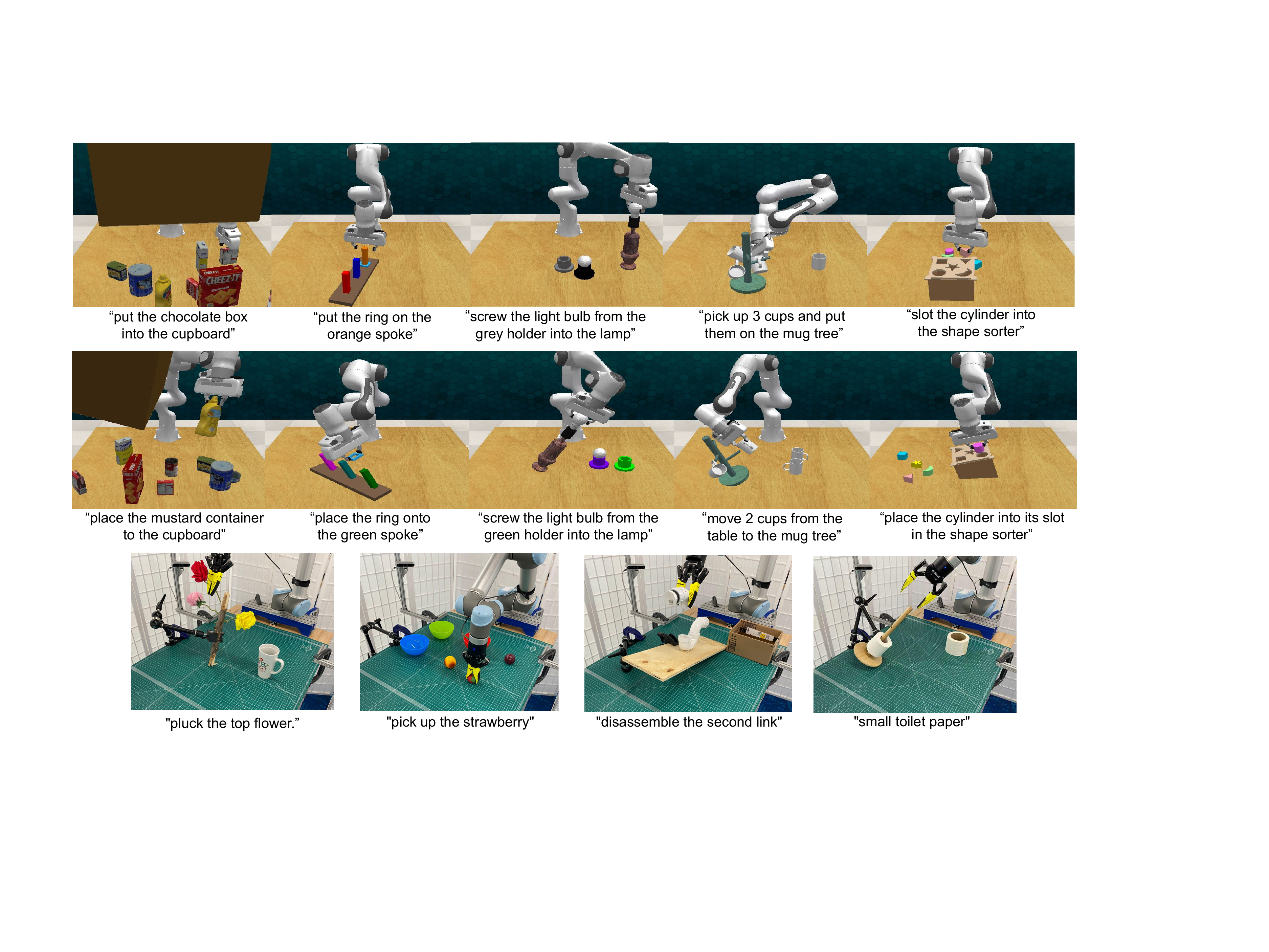}
    \vspace{-2mm}
    \caption{\textbf{Simulation and physical experiments.} First row: 18 standard RLBench tasks\cite{shridhar2023perceiver,james2020rlbench}. Second row: 18 RLBench tasks with $\SE(3)$ randomization. Third row: 4 physical experiments. A language instruction specifies each variant of the task.}
    \label{fig:exp_vis}
    \vspace{-2mm}
\end{figure}

\input{table/rlbench18}

\paragraph{Task setups.} We benchmark multi-task algorithms on 18 RLBench~\citep{shridhar2023perceiver, james2020rlbench} tasks. The benchmark uses a Franka Panda robot equipped with a parallel gripper. Observations are captured from four RGB-D cameras positioned at the front, left shoulder, right shoulder, and wrist, with resolutions of either $128^2$ or $256^2$ pixels. Each task includes several variations specified by natural language instructions. For example, in the ``open\_drawer'' task, ``open\_the\_top\_drawer'' and ``open\_the\_middle\_drawer'' are two distinct variations. Across all tasks, there are between $2$ and $60$ variations per task, resulting in a total of $249$ variations.

\textbf{Evaluation metric.} Performance is measured by a binary reward, where $0\%$ and $100\%$ correspond to failure and successful completion of the task according to the natural language instruction, respectively. We report the task success rate over 25 evaluation episodes per task, with a maximum of 25 steps per episode. During evaluation, the objects and language goals remain the same as in the training set, but the object poses are novel.

\paragraph{Baselines.} We benchmark our method with two strong baselines. \textbf{SAM2ACT}\cite{fang2025sam2actintegratingvisualfoundation} is the current state-of-the-art baseline on 18 RLBench, which leverages pretrained image tokenizer from SAM2 \cite{ravi2024sam} and projects point cloud into image planes \cite{goyal2024rvt}. \textbf{3DDA} stands for 3D diffuser actor \cite{3dda}, which takes point cloud as input and leverages diffusion policy to capture multi-modality in the demonstrations. All the baselines are trained and evaluated on a single RTX 4090 GPU with 24 GB memory. We report hyperparameters in Appendix \ref{app:hyperparameters}.

\paragraph{Experiment settings.} We benchmark baselines on three experiment settings with increasing difficulty. In the $100$ setting \cite{shridhar2023perceiver}, the model is trained with $100$ demonstrations per task, then tested with randomly $\SE(2)$ initialized objects. In the \textbf{$10$} setting, the model is trained with $10$ demonstrations per task and tested in the same way as $100$. In the $10\ \SE(3)$ setting, the training set contains $10$ demo per task and both the training and testing scenes have randomly $\SE(3)$ initialized objects.

\paragraph{Results.} Table \ref{table:rlbench18} shows that on average, EquAct outperforms all the baselines on the $100$ setting by $2.6\%$, the $10$ setting by $6.2\%$, and the $10\ \SE(3)$ setting by $15.4\%$. Furthermore, the more difficult the setting is, the more EquAct outperforms the baselines, demonstrating strong sample efficiency and 3D generalization. EquAct also excels at tasks requiring precisions, e.g., ``place\_cups'' and ``sort\_shape'', where other baselines struggle. This indicates that the equivariance is crucial for a policy adapting precisely to objects pose. Lastly, EquAct underperforms baselines in the tasks in which the object's pose is fixed, e.g., ``sweep\_to\_dustpan''. Besides success rate, EquAct matches the training/inference time and GPU memory consumption of other baselines.

\subsection{Physical experiments}
\label{sec:phy_exp}

\input{table/physcial_exp}

We benchmark the performance of EquAct and baseline on $4$ physical multi-task with $11$ variations and $135$ demonstrations in total: ``disassemble\_pipe'', ``pluck\_flower'', ``pick\_fruit'', ``install\_toilet\_roll''. The pose of objects in all tasks, except ``pick\_fruit'', undergo random SE(3) transformations within the manipulator's workspace. Details of the experiment setting are given in \autoref{sec:realworld-exp}. We evaluate 10 episodes for each task and report the binary success rate. Notice that physical settings are more challenge than simulation, due to noisy demonstrations and noisy observations. We baseline with the best model 3DDA \cite{3dda} in the 10-$\SE(3)$ setting in Table \ref{table:rlbench18}, and show quantitative results in \autoref{table:physical}. EquAct effectively learns physical SE(3) multi-task keyframe policy from limited demonstrations, achieving $65\%$ average success rate. In comparison, 3DDA struggles in these experiments, often skipping keyframe actions and resulting in failure.

\subsection{Ablation study}
\label{sec:ablation}

\input{table/ablation}

We perform the ablations on the $10$ demo setting: \textbf{Ours:} the full EquAct model. \textbf{aug. $\rightarrow$ no aug.} removes data augmentation by training with the raw demonstration data. \textbf{iFiLM $\rightarrow$ FiLM:} ablates the proposed iFiLM layers by replacing them with standard FiLM layers~\citep{perez2018film}. \textbf{$l=3\rightarrow2$:} reduces the Fourier feature resolution by restricting the spherical harmonic degree from $3$ to $2$. \textbf{equ. $\rightarrow$ no equ.:} breaks the equivariance of EquAct by replacing one equivariant graph attention layer in $q_t$ and $q_r$ with a Roformer transformer layer~\citep{su2024roformer, gervet2023act3d}.

Table~\ref{table:ablation} reports the multi-task success rates across 4 RLBench tasks. Even though only a single equivariant layer is replaced, \textbf{equ. $\rightarrow$ no equ.} results in the largest performance drop, underscoring the critical role of maintaining geometric structure in EquAct. The \textbf{$l=3\rightarrow2$} ablation highlights the importance of high-degree (high-resolution) spherical Fourier coefficients for accurate action reasoning. Additionally, while \textbf{iFiLM $\rightarrow$ FiLM} performs similarly to the full model in most cases, it shows a notable performance degradation on precision-demanding tasks such as ``place\_cups,'' demonstrating the generalization advantage of the proposed iFiLM layers. Finally, \textbf{aug. $\rightarrow$ no aug.} indicates using data augmentation can further improve performance, we hypothesize that data augmentation reduces numerical error in the equivariant neural networks.

%% file: table/rlbench18.tex
\begin{table*}[!ht]
\centering
\tiny
\hspace*{-2cm}
\caption{\textbf{Multi-task success rate (\%) on 18 RLBench.} \textbf{100} and \textbf{10} denote the number of training demonstrations per task with random $\SE(2)$ objects poses. \textbf{10*} denotes $10$ demonstrations per task with random $\SE(3)$ objects poses. On average, EquAct outperforms all the baselines on all $3$ settings.}
\begin{tabular}{cccccccccccccccc}
\toprule
       & \multicolumn{3}{c}{avg. success rate $\uparrow$} & \multicolumn{3}{c}{open drawer} & \multicolumn{3}{c}{slide block}  & \multicolumn{3}{c}{sweep dust.}  & \multicolumn{3}{c}{meat off grill}  \\ \rule{0pt}{3ex} 
Method & 10$*$ & 10 & 100 & 10$*$ & 10 & 100 & 10$*$ & 10 & 100 & 10$*$ & 10 & 100 & 10$*$ & 10 & 100  \\ \midrule
EquAct  & \textbf{53.3} & \textbf{60.1} & \textbf{89.4} & \textbf{55} & 74 & 78  & \textbf{48} & 56 & \textbf{100}  & 59 & 61 &  83 & \textbf{96} & \textbf{100}  & \textbf{100}       \\
SAM2ACT~\citep{fang2025sam2actintegratingvisualfoundation}  & 37.0 & 52.2 & 86.8  & 25 & 76 &  83  & 40 & 32  &  86 & 72 & 76 &  \textbf{99} & 80 & 72 &  98     \\
3DDA~\citep{3dda}  & 37.9 & 50.3 & 81.3  & 30 & \textbf{87} & \textbf{90}  & 43 & \textbf{72} &  98 & \textbf{95}  & \textbf{83} & 84  & \textbf{96} & 78 &  97   \\

\bottomrule \rule{0pt}{4ex} 
       & Train. & Infer. & Mem. & \multicolumn{3}{c}{screw bulb}   & \multicolumn{3}{c}{put in safe}   & \multicolumn{3}{c}{place wine}     & \multicolumn{3}{c}{put in cupboard} \\ \rule{0pt}{3ex} 
Method  & t (h) $\downarrow$  &  t (s) $\downarrow$ & (GB) $\downarrow$  & 10$*$ & 10 & 100 & 10$*$ & 10 & 100 & 10$*$ & 10 & 100 & 10$*$ & 10 & 100 \\ \midrule
EquAct  & 240 & 0.7 & 21  & \textbf{36} & 53 & 68  & \textbf{76} & \textbf{91} & \textbf{100}  & 14 & \textbf{95} & \textbf{95}  & \textbf{36} & 22 & \textbf{89}       \\
SAM2ACT~\citep{fang2025sam2actintegratingvisualfoundation}   & 225 & 0.1 & 21 & 4 & \textbf{64} & \textbf{89}    & 68 & 48 &  98 & 40 & 68 &  93 & 0 & 8 & 75      \\
3DDA~\citep{3dda}  & 253 & 3.7 & 20 & 5 & 37 & 82  & 62 & 70 & 98 & \textbf{73} & 82 & 94  & 9 &  \textbf{28} &  86    \\

\bottomrule \rule{0pt}{3ex} 
           & \multicolumn{3}{c}{close jar}  & \multicolumn{3}{c}{drag stick} & \multicolumn{3}{c}{stack blocks}   & \multicolumn{3}{c}{stack cups}  & \multicolumn{3}{c}{place cups}   \\ \rule{0pt}{3ex}
Method  & 10$*$ & 10 & 100  & 10$*$ & 10 & 100 & 10$*$ & 10 & 100 & 10$*$ & 10 & 100 & 10$*$ & 10 & 100 \\ \midrule
EquAct  & \textbf{33} & 52 & 91 & \textbf{85} & 90 &  95 & \textbf{26} & \textbf{35} & \textbf{90}  & \textbf{59} & \textbf{18} & 68  & \textbf{21} & \textbf{62} & \textbf{76}   \\
SAM2ACT~\citep{fang2025sam2actintegratingvisualfoundation}  & 8 &  \textbf{68} & \textbf{99} & 44 & \textbf{100} &  99  & 16 & 20 & 76 & 0 & 12 & \textbf{78}  & 0 & 4 & 47     \\
3DDA~\citep{3dda}  & 24 & 52 & 96  & 60 & 35 & \textbf{100}  & 16 & 10 & 68 & 9 & \textbf{18} & 47 & 0 & 10 & 24    \\

\bottomrule \rule{0pt}{3ex} 
            & \multicolumn{3}{c}{turn tap}   & \multicolumn{3}{c}{put in drawer} & \multicolumn{3}{c}{sort shape} & \multicolumn{3}{c}{push buttons}   & \multicolumn{3}{c}{insert peg}  \\ \rule{0pt}{3ex}
Method  & 10$*$ & 10 & 100  & 10$*$ & 10 & 100 & 10$*$ & 10 & 100 & 10$*$ & 10 & 100 & 10$*$ & 10 & 100 \\ \midrule
EquAct  & \textbf{67} & 56 & \textbf{100} & \textbf{64} & 84 & \textbf{100}  & \textbf{36} & \textbf{33} & \textbf{86}  & \textbf{89} & 85 & \textbf{100}  & \textbf{60} & 14 & \textbf{90}       \\
SAM2ACT~\citep{fang2025sam2actintegratingvisualfoundation}  & 36 & \textbf{92} & 96 & \textbf{64} & \textbf{100} &  99   & 24 & 16 & 64  & 40 & 56 & \textbf{100}  & 4 & \textbf{28}  &  84    \\
3DDA~\citep{3dda}  & 48 & 74 & 99  & 14 & 92 & 96  & 14 & \textbf{33} & 44  & 79 & 95 &   98 & 5 & 14 & 66    \\
\bottomrule
\end{tabular}
\label{table:rlbench18}
\end{table*}

%% file: table/physcial_exp.tex
\begin{wraptable}{r}{0.55\textwidth}
\vspace*{-1cm}
\centering
\caption{\textbf{Physical experiments.}}
\scriptsize
\begin{tabular}{cccccc}
\toprule
       & avg.           & disass.   & pluck   & pick & install       \\
       & SR $\uparrow$  & pipe          & flower    & fruit & toilet roll   \\
       Var $\times$ Demo & & $3\times10$ & $3\times15$ & $3\times10$ & $2\times15$ \\\midrule
Ours  & 65.0 & 90  & 70  & 50  & 50 \\
3DDA\cite{3dda}   &12.5 &0 &20 & 30& 0 \\ 
\bottomrule
\end{tabular}
\label{table:physical}
\vspace*{-0.2cm}
\end{wraptable}


%% file: table/ablation.tex
\begin{wraptable}{r}{0.38\textwidth}
\vspace{-1cm}
\setlength\tabcolsep{2.3pt}
\caption{\textbf{Ablation study.}}
\scriptsize
\begin{tabular}{cccccccccccccccc}
\toprule
       & avg.            & place   & place  & reach & insert       \\
       & SR $\uparrow$   & wine    & cups   & drag  & peg           \\ \midrule
Ours   & 52.8 & 45 & 62 & 90 & 14            \\
aug. $\rightarrow$ no aug.   & 50.5 & 36 & 71 & 85 & 10           \\
iFiLM $\rightarrow$ FiLM & 50.3 & 68 & 24 & 90 & 19        \\
$l=3\rightarrow2$   & 45.5 & 64 & 28 & 80 & 10          \\
equ. $\rightarrow$ no equ.   & 12.3 & 14 & 0 & 35 & 0      \\
\bottomrule
\end{tabular}
\label{table:ablation}
\end{wraptable}

%% file: section/6_conclusion.tex
\section{Conclusion and limitations}
\label{sec:conclusion_limitation}

\paragraph{Conclusion.} This paper proposes EquAct to leverage $\SE(3)$ equivariance in the multi-task keyframe policy and invariance in the language instruction. Specifically we use a novel equivariant point transformer U-net (EPTU) to encode the observation and use equivariant field networks to evaluate action candidates. Then we propose invariant FiLM layers to modulate the policy with natural language instructions. In the end, EquAct outperforms SOTA baselines by $2.6\%$ and $6.2\%$ when trained with $100$ or $10$ demos in $\SE(2)$ setting, and by $15.4\%$ when trained with $10$ demos in $\SE(3)$ setting. Physical experiments validated that EquAct can solve complex tasks with $\SE(3)$ variation.

\paragraph{Limitations.} The keyframe action formulation assumes the task can be solved by several key gripper poses. This assumption is satisfied in RLBench tasks but could be broken in the closed-loop settings \cite{chi2023diffusion}. 
Moreover, despite EquAct scales well with training data in Table \ref{table:rlbench18}, the data efficiency and semantic generalization could be further improved by leveraging pre-trained vision models \cite{radford2021learning,shafiullah2022clip,gervet2023act3d}. Lastly, the training and the inference speed of EquAct is slower than the best baseline; a more efficient equivariant backbone can speed up the inference.

%% file: section/7_appendix.tex
\newpage

\section{Proofs}
\label{app:proofs}

\subsection{Proof of proposition \ref{proposition:equact}:}
\label{proof:equact}
\begin{proof}
To prove the equivariance of EquAct with respect to $o$, we only need to prove that every layer of EquAct is equivariant, then by induction, EquAct is equivariant to the observation $o$. See Proof \ref{proof:smaxpool}, \ref{proof:sup} that proves the equivariance of the proposed spherical maxpool layers and the proposed spherical upsampling layers. Referring ~\citep{liaoequiformerv2} for proof of Equiformer layers and ~\citep{cohen2018spherical} for proof of Spherical CNNs.

To prove the invariance of EquAct with respect to the nature language instruction $n$, we only need to prove that the iFiLM layers are invariance to $n$, see Proof \ref{proof:iFiLM}.
\end{proof}

\subsection{Proof of proposition \ref{proposition:smaxpool}:}
\begin{proof}
\label{proof:smaxpool}
Focusing on the right-hand side of Equation ~\ref{pro:smaxpool}, and denoting the point with the largest manganite of Fourier coefficients after transformation $g=r \ltimes t$ as $p^*_g$:
\begin{equation}
    \text{smaxpool}\{ \mathrm{D}(r) \cdot c_{l,p} | p\in t + knn(x) \} = \mathrm{D}(r) \cdot c_{l,p^*_g}\\
\end{equation}
Expanding the equation of $p^*_g$, and using the property that the winger-D matrices are orthogonal, we have:
\begin{align}
    p^*_g &= {\arg\max}_{p \in t + knn(x)} \| \mathrm{D}(r) \cdot c_{l,p} \|_2^2 \\
    &= {\arg\max}_{p \in t + knn(x)}  \big((\mathrm{D}(r) \cdot c_{l,p})^\text{T}(\mathrm{D}(r) \cdot c_{l,p})\big) \\
    &= {\arg\max}_{p \in t + knn(x)}  (c_{l,p}^\text{T} \cdot \mathrm{D}(r)^\text{T}\mathrm{D}(r) \cdot c_{l,p}) \\
    &= {\arg\max}_{p \in t + knn(x)}  (c_{l,p}^\text{T}c_{l,p}) \\
    &= {\arg\max}_{p \in t + knn(x)}  \| c_{l,p} \|_2^2 \\
    &= t + {\arg\max}_{p \in knn(x)}  \| c_{l,p} \|_2^2 \\
    &= t + p^*
\end{align}
Thus:
\begin{equation}
    \text{smaxpool}\{ \mathrm{D}(r) \cdot c_{l,p} | p\in t + knn(x) \} = \mathrm{D}(r) \cdot c_{l,p^*_g} = \mathrm{D}(r) \cdot c_{l,t + p^*} = \mathrm{D}(r) \cdot c_{l,\, t + x}'
\end{equation}
\end{proof}

\subsection{Proof of proposition \ref{proposition:sup}:}
\begin{proof}
\label{proof:sup}
Expanding the right-hand side of Equation \ref{pro:sup} gives:
\begin{align}
    \text{sup}\{\mathrm{D}(r) \cdot c_{l,\, p'},\, t + x | p'\in t+\text{knn}(x)\} &= \text{softmax}_{p'\in t+\text{knn}(x)}\left( \frac{1}{\|t + x - p'\|} \right) \mathrm{D}(r) \cdot c_{l,p'} \\
    &= \text{softmax}_{t + p\in t+\text{knn}(x)}\left( \frac{1}{\|t + x - t - p\|} \right) \mathrm{D}(r) \cdot c_{l,t + p} \\
    &= \text{softmax}_{p\in \text{knn}(x)}\left( \frac{1}{\|x - p\|} \right) \mathrm{D}(r) \cdot c_{l,t + p} \\
    &= \mathrm{D}(r) \cdot\text{softmax}_{p\in \text{knn}(x)}\left( \frac{1}{\|x - p\|} \right)  c_{l,t + p} \label{line:linear}\\
    &= \mathrm{D}(r) \cdot c_{l,t + x}'
\end{align}
Line \ref{line:linear} is due to Schur’s lemma~\citep{schur1905neue}, which proved that any linear combination of Fourier coefficients is equivariant.
\end{proof}

\subsection{Proof of proposition \ref{proposition:iFiLM}:}
\begin{proof}
\label{proof:iFiLM}
When $l=0$, the wingle-D matrix is an identity matrix, thus:
\begin{equation}
    \beta \big(\mathrm{D}(r) \cdot c_0\big) + \gamma = \beta c_0 + \gamma = c_0' = \mathrm{D}(r) \cdot  c_0'
\end{equation}
When $l>0$, expanding the right-hand side of Equation \ref{line:l>0} and applying Schur's lemma \citep{schur1905neue} we have:
\begin{align}
    \alpha_l \big(\mathrm{D}(r) \cdot c_l\big) = \mathrm{D}(r) \cdot (\alpha_l c_l) = \mathrm{D}(r) \cdot c_l'
\end{align}
\end{proof}

\section{$18$ RLBench tasks with standard and $\SE(3)$ initializations}

The $18$ RLBench tasks \cite{shridhar2023perceiver,james2020rlbench} are initialized with objects in random $\SE(2)$ poses. In this paper, we present $18$ RLBench tasks with $\SE(3)$ variation, where in addition to the $\SE(2)$ initialization, the pose of objects are further perturbed with $\SO(3)$ transformation. This change will leads keyframe actions change in $\SE(3)$. For detailed $\SO(3)$ perturbation range and perturbed object, see \autoref{table:rlbench_se3}.

\begin{table}[!h]
\centering
\setlength\tabcolsep{2.3pt}
\caption{\textbf{18 Language-conditioned tasks in RLBench}~\citep{james2020rlbench} with SE(3) initializations.}
\scriptsize
\begin{tabular}{llccl} 
\toprule
Task                      & Variation   Type           & Perturbed Object     & $\SO(3)$ Perturbation ($r,p$)      & Language     Template         \\
\midrule
\texttt{open drawer}      & placement                  & \texttt{drawer}                &   
$[0, -0.5], [0.6, 0.5]$         & ``open the \blank drawer'' \\
\texttt{slide block}      & color                      & \texttt{plane}              &    $[-0.12, -0.12], [0.12, 0.12]$               & ``slide the block to \blank target'' \\
\texttt{sweep to dustpan} & size                       &  \texttt{broom holder}         &     $ [0, 0, -0.9], [0, 0, 0.9]$              & ``sweep dirt to the \blank dustpan'' \\
\texttt{meat off grill}   & category                   &          \texttt{grill table}           &     $ [-0.25, -0.25], [0.25, 0.25]$             & ``take the \blank off the grill'' \\
\texttt{turn tap}         & placement                  &           \texttt{tap}         &      $ [-0.5, -0.5], [0.5, 0.5]$               & ``turn \blank tap'' \\
\texttt{put in drawer}    & placement                  &           \texttt{drawer}         &      $ [0, -0.2], [0.2, 0.2]$            & ``put the item in the \blank drawer'' \\
\texttt{close jar}        & color                      &         \texttt{jar}         &       $ [0, -0.5], [0.6, 0.5]$            & ``close the \blank jar'' \\
\texttt{drag stick}       & color                      &          \texttt{plane}          &     $ [-0.12, -0.12], [0.12, 0.12]$               & ``use the stick to drag the cube onto the \blank target'' \\
\texttt{stack blocks}     & color, count               &          \texttt{plane}          &       $ [-0.15, -0.15], [0.15, 0.15]$            & ``stack \blank \blank blocks''  \\
\texttt{screw bulb}       & color                      &         \texttt{lamp base}         &    $[-0.6, -0.6], [0.6, 0.6]$             & ``screw in the \blank light bulb'' \\
\texttt{put in safe}      & placement                  &           \texttt{safe}          &     $[-0.25, -0.3], [0.5, 0.3]$             & ``put the money away in the safe on the \blank shelf'' \\
\texttt{place wine}       & placement                  &          \texttt{wine rack}         &     $[-0.5, -0.5], [0.5, 0.5]$               & ``stack the wine bottle to the \blank of the rack'' \\
\texttt{put in cupboard}  & category                   &           \texttt{cupboard}         &   $[-0.5, -0.5], [0.5, 0.5]$           & ``put the \blank in the cupboard'' \\
\texttt{sort shape}       & shape                      &           \texttt{shape sorter}         &   $[-0.25, -0.25], [0.25, 0.25]$                  & ``put the \blank in the shape sorter'' \\
\texttt{push buttons}     & color                      &         \texttt{buttons}          &   $[-0.25, -0.25], [0.25, 0.25]$                & ``push the \blank button, [then the \blank button]'' \\
\texttt{insert peg}       & color                      &          \texttt{pillars}          &     $[-0.3, -0.4], [0.3, 0.4]$               & ``put the ring on the \blank spoke'' \\
\texttt{stack cups}       & color                      &          \texttt{cups}         &    $[-0.3, -0.3], [0.3, 0.3]$               & ``stack the other cups on top of the \blank cup'' \\
\texttt{place cups}       & count                      &          \texttt{cups}        &   $[0, -0.5], [0.6, 0.5]$           & ``place \blank cups on the cup holder'' \\
\bottomrule
\end{tabular}
\label{table:rlbench_se3}
\end{table}

%

\section{Details of $4$ physical tasks}
\label{sec:realworld-exp}

\begin{figure}[h]
\centering
\includegraphics[width=0.9\linewidth]{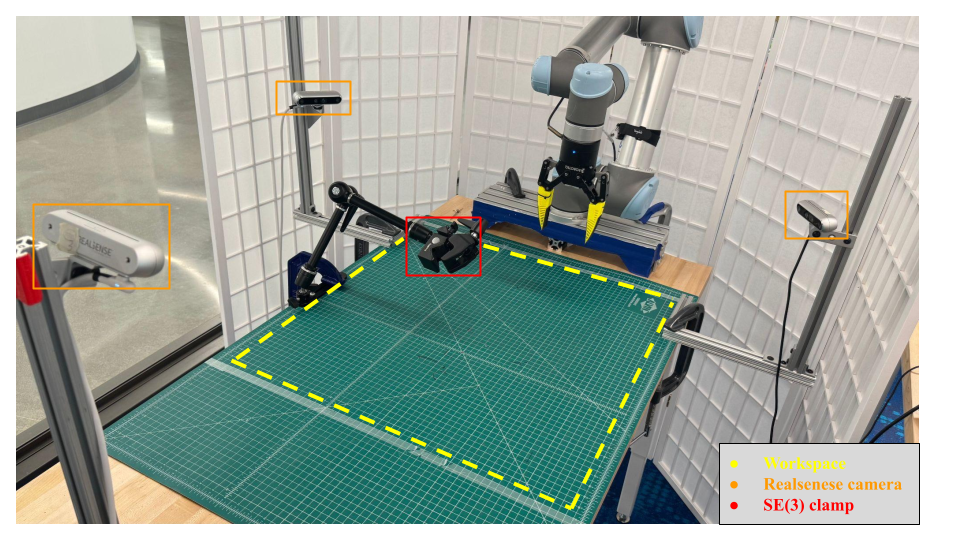}
\caption{\textbf{Real world experimental setup}}
\label{fig:realworld_setting} 
\end{figure}

Our real-world experiments are carried out on a UR5 robotic arm equipped with a Robotiq 2F-85 gripper and three Intel RealSense D455 cameras, as shown in \autoref{fig:realworld_setting}. Keyframe actions are collected using a 6-DoF 3DConnexion SpaceMouse, collecting both visual observations (from all three cameras) and robot end-effector actions (position, orientation, and gripper states).

\begin{figure}[!h]
\centering
\includegraphics[width=1\linewidth]{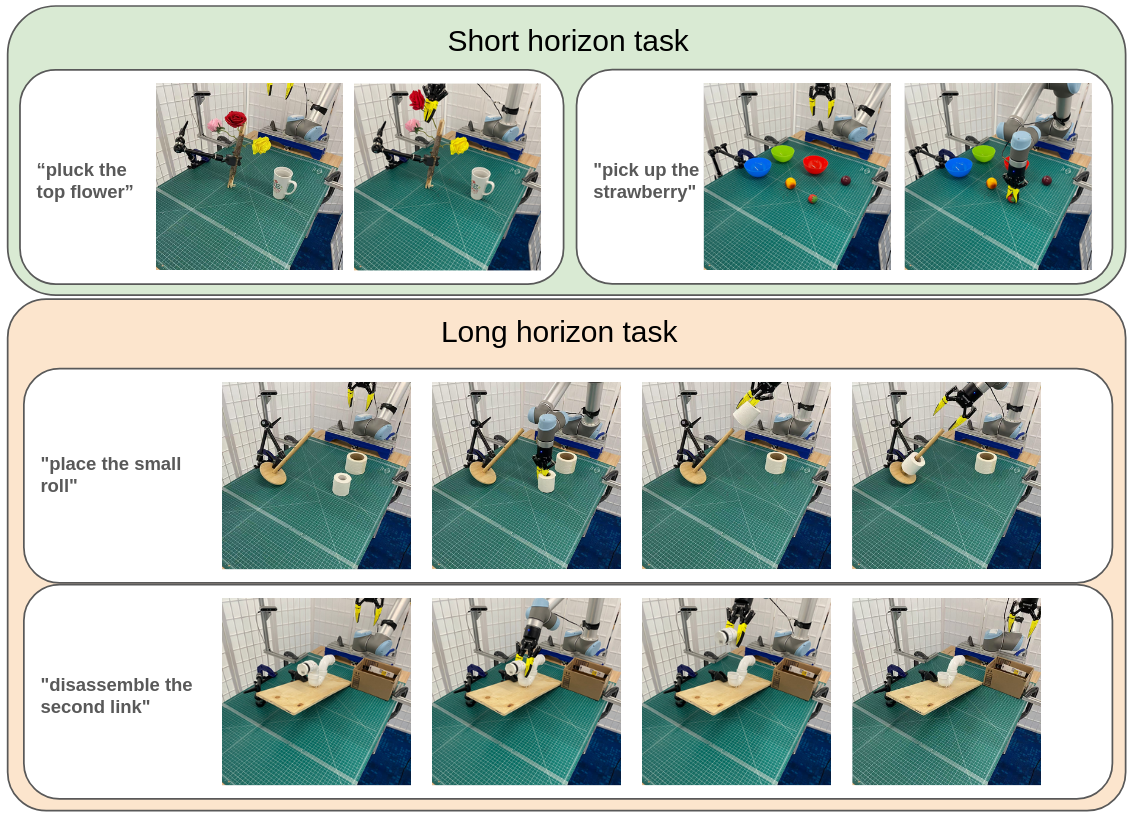}
\caption{\textbf{$4$ Physical tasks.}}
\label{fig:physical_tasks} 
\end{figure}

The $4$ physical tasks are visualized in \autoref{fig:physical_tasks}. For details of these tasks, see descriptions below.

\subsection{Disassemble pipe}

\textbf{Task:} Disassemble the required link of the pipe: first, second, all. 

\textbf{Number of keyframe actions:} 3-10.

\textbf{Variations:} "first link", "second link","all".

\textbf{Objects:} Pipes consisting of five sections of water pipes.

\textbf{Success Metric}: The robot must accurately grasp the target pipe segment and completely remove it from the intact assembly. 

\subsection{Pluck flowers}

\textbf{Task:} Pluck the specified flower: top, middle, bottom.

\textbf{Number of keyframe actions:} 4.

\textbf{Variations:} "top flower", "middle flower", "bottom flower".

\textbf{Objects:} Three artificial flowers and one vase.

\textbf{Success Metric:} The robot must accurately grab the designated flower and pluck it.

\subsection{Pick fruit}

\textbf{Task:} Pick up the specified
fruit(strawberry,peach,plum).

\textbf{Number of keyframe actions:} 3.

\textbf{Variations:} "strawberry", "peach", "plum".

\textbf{Objects:} Three fruits of mixed types.

\textbf{Success Metric:} The robot must correctly identify, grasp the target fruit.

\subsection{Install toilet roll}

\textbf{Task:} Place the specified toilet paper roll: large, small.

\textbf{Number of keyframe actions:} 5.

\textbf{Variations:} "large roll", "small roll".

\textbf{Objects:} Two toilet-paper rolls and one wall-mounted holder.

\textbf{Success Metric:} The robot must pick up the specified roll and mount it onto the holder.

\section{Hyperparameters}
\label{app:hyperparameters}
We report the following hyperparameters in \autoref{table:hyperparameters} for EquAct as well as baselines we compared in the paper.

\begin{table}[!h]
\centering
\setlength\tabcolsep{2.3pt}
\caption{\textbf{Hyperparameters.} sim: the hyperparameters used in simulation experiments. phy: the hyperparameters used in physical experiments.}
\scriptsize
\begin{tabular}{lccccc} 
\toprule
Name of the      &  \multicolumn{5}{c}{Method}        \\
hyperparameter & EquAct (sim) & EquAct (phy) & 3DDA (sim) & 3DDA (phy) & SAM2ACT (sim) \\
\midrule
\# $a_t$ (train/test)  & 450/3000 & 450/6000 & None & None & None \\
learning rate & 1e-4 & 1e-4 & 1e-4 & 1e-4 & 1e-4 \\
lr scheduler & None & None & None & None & cosine \\
batch size & 2 & 2 & 7 & 7 & 8 \\
training iterations & 8e5 & 6e4 & 6e5 & 1.2e5 & 5.625e4 \\
\bottomrule
\end{tabular}
\label{table:hyperparameters}
\end{table}